%% file: 083_main.tex
\documentclass[runningheads]{llncs}
\usepackage[compatibility=false]{caption}
\usepackage{lmodern}
\usepackage{textcomp}
\RequirePackage{silence}  
\usepackage{graphicx}

\newcommand\samethanks[1][\value{footnote}]{\footnotemark[#1]}
\usepackage{todonotes}
\usepackage{amsmath}
\usepackage{booktabs}
\usepackage{amssymb}   
\usepackage{enumitem}
\usepackage{orcidlink}
\usepackage{subcaption}
\usepackage[percent]{overpic}
\usepackage[table]{xcolor}
\usepackage{makecell}
\newcommand{\Lklsym}{\mathcal{L}_{\mathrm{KL\text{-}sym}}}
\newcommand{\Lklasym}{\mathcal{L}_{\mathrm{KL\text{-}asym}}}
\newcommand{\cw}[1]{\textcolor{red}{#1}}          
\newcommand{\cb}[1]{\textcolor{green!55!black}{#1}} 
\newcommand{\mc}[2]{\makecell{#1\\[-1pt]{\scriptsize #2}}} 

\newcommand{\ms}[2]{#1\,{\scriptsize$\pm#2$}}
\usepackage{multirow}

\usepackage{comment}
\usepackage{amsmath,amssymb}
\usepackage{color}
\usepackage{url}
\usepackage{hyperref}

\newcommand{\model}{CRAFT}

\newif\ifreview
\reviewtrue
\reviewfalse

\ifreview
	\usepackage{lineno}

	\linenumbers
\fi

\begin{document}


\def\SubNumber{83}

\def\GCPRTrack{Fast Review Track}

\title{Learning Where to Focus: Self-Supervised Multi-Scale ViTs for Histopathology}

\ifreview
	\titlerunning{Learning Where to Focus}
	\authorrunning{GCPR 2026 Submission \SubNumber{}. CONFIDENTIAL REVIEW COPY.}
	\author{GCPR 2026 - \GCPRTrack{}}
	\institute{Paper ID \SubNumber}
\else
	\titlerunning{Learning Where to Focus}

	\author{Anabel Stammer\thanks{Corresponding Author}\orcidID{0009-0009-4885-0630} \and
	Valay Bundele\thanks{These authors contributed equally.}\orcidID{0000-0003-2140-9019} \and
	Mehran Hosseinzadeh\samethanks\orcidID{0009-0000-1114-1595} \and
    Hendrik P.A. Lensch\orcidID{0000-0003-3616-8668}}
	
	\authorrunning{A. Stammer et al.}
	

    \institute{Eberhard Karls Universität Tübingen, Germany
    \email{\{anabel.stammer,valay.bundele,mehran.hosseinzadeh, hendrik.lensch\}@uni-tuebingen.de}}
\fi

\maketitle              

\begin{abstract}
Pathologists diagnose diseases by first locating suspicious tissue and then examining it at higher magnification, whereas self-supervised vision transformers (ViTs) allocate the same spatial resolution to every image region despite diagnostic evidence being sparse and spanning multiple biological scales. Recent pathology foundation models have substantially improved representation quality by scaling training data and model capacity, but largely retain uniform tokenization. We instead investigate whether pathology representations can be improved by learning where to allocate spatial resolution during self-supervised learning. To this end, we propose CRAFT (Coarse-to-fine Region-Adaptive Feature Tokenization), a DINO-based framework that learns image-dependent mixed-scale representations by using self-supervised attention to selectively refine informative regions while preserving coarse context, together with a symmetric cross-scale regularization objective that encourages complementary coarse and fine representations. Across CAMELYON16, TCGA-Lung subtype classification, and TCGA-LUAD survival prediction, CRAFT consistently outperforms comparable-scale self-supervised methods while requiring lower inference computation. Despite using only a compact 22M parameter backbone trained on comparatively small pathology datasets, CRAFT remains competitive with, and often surpasses, substantially larger pathology foundation models.

\end{abstract}
\section{Introduction}
\label{sec:intro}
Pathology is as much a search problem as a recognition problem: in whole-slide
images (WSIs), diagnostically relevant tissue is sparse and spans multiple
biological scales~\cite{chen2022HIPT}. Pathologists navigate this by scanning the
slide broadly, then examining only suspicious regions at higher magnification. Most vision transformers (ViTs) for computational pathology instead tokenize every region at a single fixed scale, forcing one representation to capture both global context and fine cellular detail. \\
Recently, self-supervised learning (SSL) has substantially advanced pathology representation learning, leading to foundation models such as UNI~\cite{chen2023UNI}, Virchow~\cite{vorontsov2024virchow}, and Prov-GigaPath~\cite{Xu2024}. Their success is largely driven by scaling training data and model capacity while retaining uniform tokenization. We instead ask a complementary question: \textit{can representation quality be improved by learning where to allocate spatial resolution during self-supervised learning?}
\begin{figure}[t]
    \centering
    \begin{overpic}[width=0.8\textwidth]{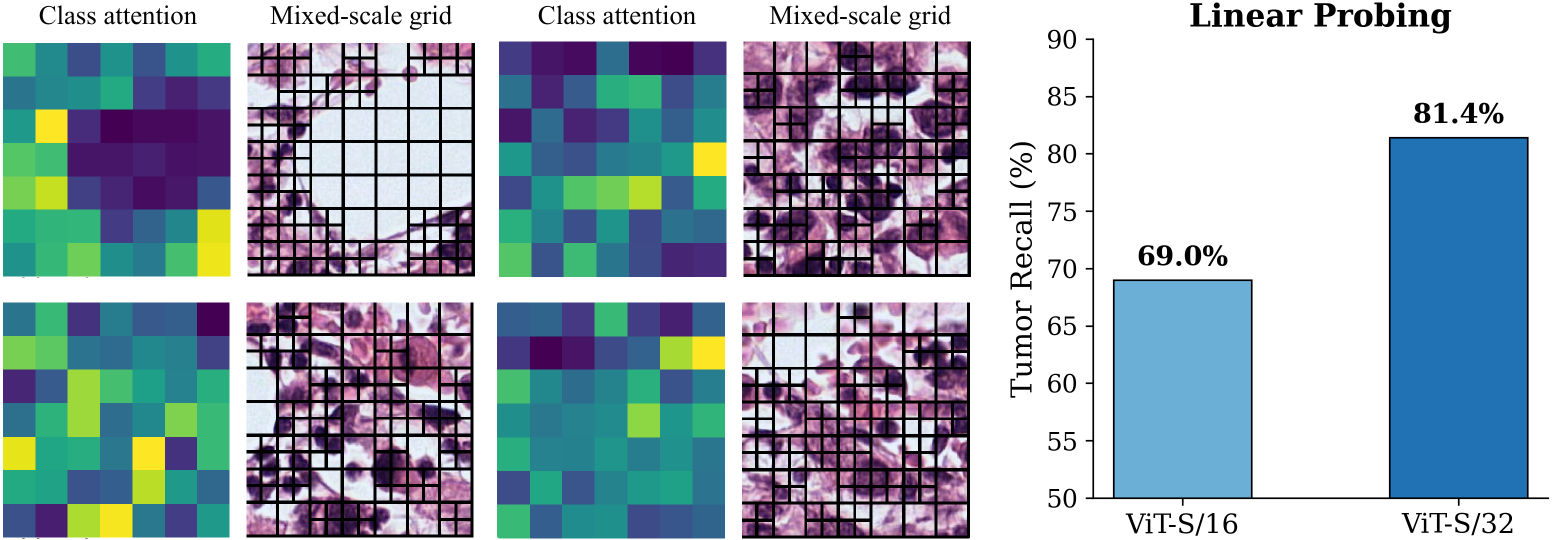}
        \put(20,-3.5){\textbf{(a)}}   
        \put(80,-3.5){\textbf{(b)}}   
    \end{overpic}
    \vspace{1.5mm}
    \caption{\textbf{(a)} Class attention maps guide mixed-scale tokenization by refining salient regions while keeping others coarse. \textbf{(b)} Tumor recall via linear probing on CAMELYON16 after SSL pretraining at two token sizes (16 vs 32).}
    \label{fig:teaser}
\end{figure}

Our motivation comes from a simple but surprising observation. Uniformly finer tokenization is not necessarily beneficial for pathology. As shown in Fig.~\ref{fig:teaser}(b), on CAMELYON16~\cite{bejnordi2017camelyon}, a DINO-pretrained ViT using coarse $32\times32$ tokens achieves substantially higher tumor recall under linear probing than the same architecture pretrained with uniformly finer $16\times16$ tokens (81.4\% vs.\ 69.0\%). This does not suggest that cellular detail is unimportant. Rather, it suggests that much of the diagnostic signal is already captured at coarse resolution, while allocating fine resolution uniformly can spend representational capacity where coarse context is sufficient. The appropriate inductive bias is therefore not uniformly coarse or uniformly fine, \textit{but coarse first, fine where needed.}

Learning such adaptive resolution during self-supervised pretraining is challenging. Without patch-level annotations, \textit{the model must first determine which regions deserve refinement}. Moreover, once coarse and refined representations coexist, \textit{the refined representation must contribute complementary information} rather than simply reproduce the coarse one. Both problems must be solved for adaptive tokenization to improve representation quality.

Existing methods address only parts of this problem. Token reduction methods such as ToMe~\cite{bolya2022token}, EViT~\cite{liang2022not}, STELLAR~\cite{zhao2026learningsparsevisualrepresentations}, and SimPrune~\cite{simprune} reduce computation by pruning or merging tokens after they are formed. WSI-specific methods typically improve efficiency after tile features have been extracted through instance selection, sparse aggregation, or redesigned Multiple Instance Learning (MIL) pipelines~\cite{araujo2024keypatchesneedmultiple,stegmuller2022scorenetlearningnonuniformattention,jafarinia2024snuffy,yang2022remixgeneralefficientframework,rahman2026dtcwsi}. Multi-scale pathology encoders such as HIPT~\cite{chen2022HIPT} and PLUTO~\cite{juyal2024plutopathologyuniversaltransformer} use predefined hierarchies rather than adapting resolution to each input, while CF-ViT~\cite{chen2023cf} explores coarse-to-fine tokenization for natural images using supervised region labels that are rarely available in histopathology. Consequently, existing methods do not learn where additional spatial resolution should be allocated during self-supervised representation learning.

We address this challenge with \textbf{CRAFT} (\textbf{C}oarse-to-fine \textbf{R}egion-\textbf{A}daptive \textbf{F}eature \textbf{T}okenization), a DINO-based self-supervised framework that learns mixed-resolution representations without region supervision. CRAFT first constructs a coarse representation of the image and uses coarse-stage class attention as a label-free saliency signal to identify informative regions. Only these regions are re-tokenized at finer resolution, while the remaining regions retain their coarse representation (Fig.~\ref{fig:teaser}(a)). The refined features are fused with their parent coarse features, allowing local morphology to complement broader tissue context. Crucially, refinement is not heuristic nor supervised: DINO’s cross-view alignment objective implicitly drives the model to allocate higher resolution to regions that most reduce representation discrepancy across views. Moreover, to ensure refinement contributes genuinely new information, CRAFT further introduces a symmetric cross-scale KL regularizer that aligns coarse and refined predictions while discouraging redundant representations.

On CAMELYON16, TCGA-Lung subtyping, and TCGA-LUAD survival prediction, CRAFT outperforms comparable-scale methods while requiring lower inference compute. Despite using only a compact 22M-parameter backbone trained on comparatively small datasets, CRAFT achieves performance competitive with large-architecture pathology foundation models trained on substantially larger datasets. These results demonstrate that learning where to allocate spatial resolution during SSL pretraining yields more effective representations than uniformly processing all WSI regions.

\section{Related Works}
\label{sec:relworks}
\subsubsection{Self-Supervised Learning in Histopathology.}

Self-supervised learning has become the dominant paradigm for learning pathology representations from large unlabeled WSI data, where dense expert annotation is impractical. Contrastive learning~\cite{ciga2022self,wessels2023self}, self-distillation~\cite{campanella2023computational,nasiri2024vim4path}, and masked-image modeling~\cite{wang2023pyramid,filiot2023scaling} have demonstrated strong transfer across downstream pathology tasks. More recently, pathology foundation models such as UNI~\cite{chen2023UNI} and Virchow~\cite{vorontsov2024virchow} have shown that scaling DINO-style pretraining with larger datasets and backbones further improves representation quality. Despite these advances, existing SSL methods uniformly tokenize every image region. In contrast, CRAFT learns \emph{where} to allocate spatial resolution during self-supervised representation learning.

\subsubsection{Efficient and Adaptive Vision Transformers.}

Efficient Vision Transformers reduce computation by adaptively processing the token sequence. Representative approaches prune, merge, or sample informative tokens~\cite{liang2022not,fayyaz2022adaptive,kong2022spvit,bolya2022token,zong2022self,simprune}, while others introduce hierarchical or coarse-to-fine representations~\cite{tang2022quadtree,havtorn2023msvit,chen2023cf}. Although effective for natural-image recognition, these methods either discard tokens or rely on supervised guidance for adaptive refinement. CRAFT instead performs region selection using self-supervised class attention and learns coarse and refined representations through cross-scale regularization, enabling adaptive spatial allocation without region-level supervision.

\subsubsection{Multi-Scale and Efficient WSI Modeling.}

The hierarchical organization of tissue has motivated multi-scale pathology models such as HIPT~\cite{chen2022HIPT} and PLUTO~\cite{juyal2024plutopathologyuniversaltransformer}, which explicitly model information across magnification levels. Separately, efficient WSI analysis has been explored through instance selection, sparse aggregation, zooming strategies, and feature compression within multiple-instance learning pipelines~\cite{ilse2018attentionbaseddeepmultipleinstance,2021joint_spatial_magnification,whole_slide_reinforcement,thandiackal2022differentiablezoomingmultipleinstance,araujo2024keypatchesneedmultiple,stegmuller2022scorenetlearningnonuniformattention,jafarinia2024snuffy,yang2022remixgeneralefficientframework,rahman2026dtcwsi}. These approaches primarily operate after patch representations have been extracted or rely on predefined multi-scale hierarchies. In contrast, CRAFT learns image-dependent resolution allocation during patch-level self-supervised learning, producing unified mixed-scale representations before slide-level aggregation.

\section{Methodology}
\label{sec:method}
\subsubsection{Preliminaries.}
DINO~\cite{caron2021emerging} trains a student network to match a stop-gradient teacher over multiple augmented views of the same image. The student receives global and local crops, the teacher receives global crops, and both networks map the \texttt{[CLS]} token through a projection head to a probability distribution over prototypes. The teacher is an exponential moving average of the student, with centering and sharpening used to avoid collapse. We build on DINO because local-to-global self-distillation naturally couples fine evidence with broader context, matching the hierarchical structure of histopathology~\cite{chen2022self_vit_path}.

\subsubsection{CRAFT.}
\begin{figure}[t]
  \centering
  \begin{subfigure}[b]{\textwidth}
    \centering
    \includegraphics[width=0.95\textwidth]{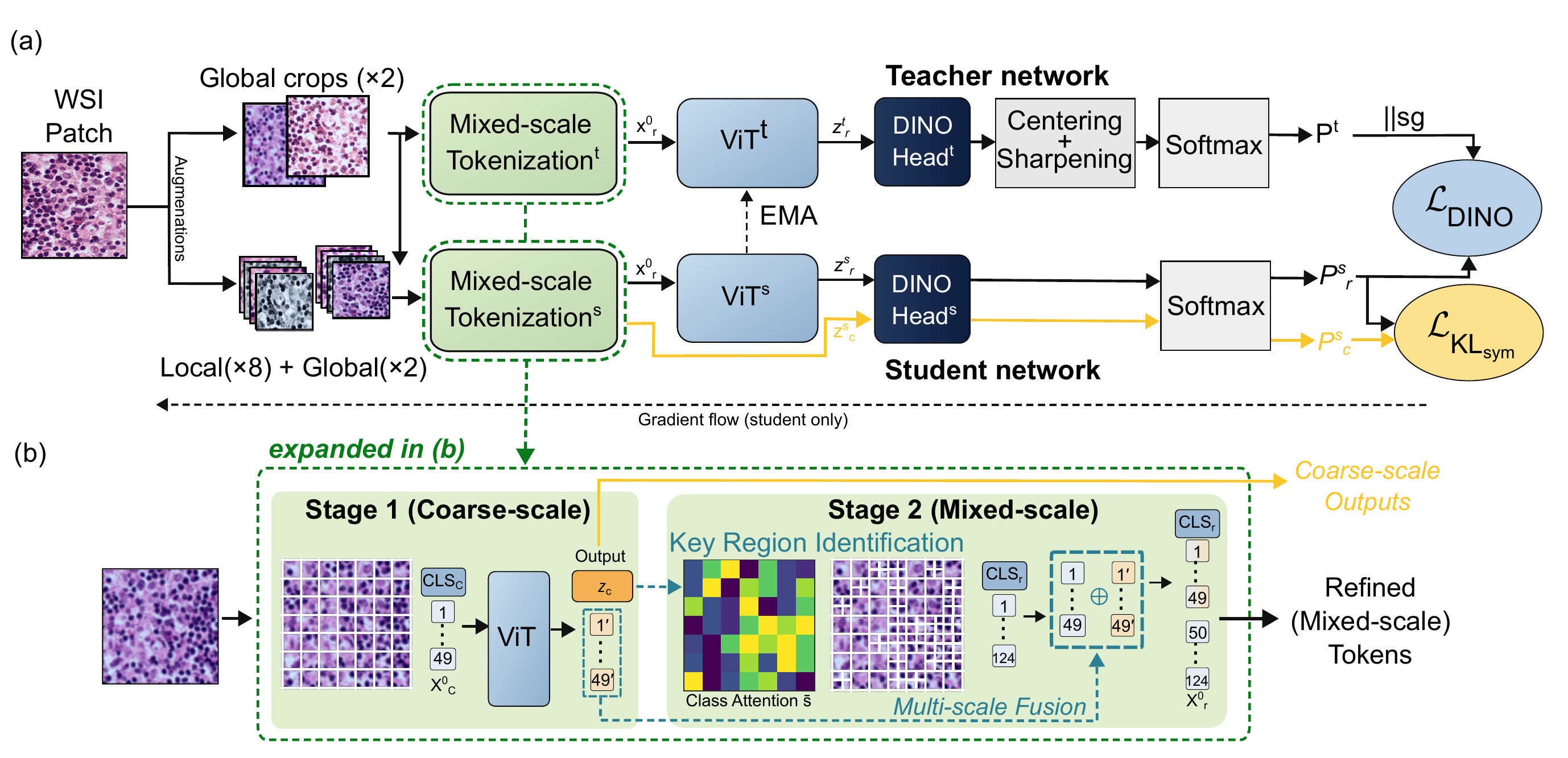}
    \label{fig:arch-dino}
  \end{subfigure}
  \caption{\textbf{(a)} Mixed-scale tokenization (green, \emph{Ours}) is inserted into both DINO branches before the ViT; the rest is standard DINO, with gradients through the student only. Coarse and refined predictions $P_c^s$, $P_r^s$ are aligned by $\mathcal{L}_{\mathrm{KL\text{-}sym}}$ (Eq.~\ref{eq:kl_sym}). \textbf{(b)} Coarse-to-fine pipeline: Stage~1 encodes coarse tokens and selects key regions by class attention; Stage~2 re-splits and fuses them ($\oplus$).}

\label{fig:architecture}
\end{figure}
\model{} turns the coarse-first principle into a self-supervised tokenization module. Instead of assigning one patch size to every image region, CRAFT forms a full-coverage mixed-scale sequence: a coarse pass preserves global tissue context, and a refined pass spends fine tokens only on regions selected from the coarse representation. The mixed-scale tokenization module is inserted symmetrically in the student and teacher branches of DINO (Fig.~\ref{fig:architecture}a), so resolution allocation is learned during feature formation rather than applied as post-hoc compression.

Unlike CF-ViT~\cite{chen2023cf}, which uses label supervision to guide
coarse-to-fine refinement, CRAFT learns region selection directly from
self-supervised pre-training. This introduces a constraint single-network
methods do not face: the tokenizer must act in both branches, and student and
teacher must select consistently or the distillation target becomes unstable.
We find the two branches converge to near-identical selections without an
explicit cross-branch mechanism (App.~I), unlike single-branch pruning
schemes~\cite{simprune}. Refinement is further optimized jointly with DINO
through a novel symmetric KL objective encouraging refined representations to
capture complementary information while remaining semantically aligned with
their coarse counterparts.

\subsubsection{Mixed-scale Tokenization.}
As shown in Fig.~\ref{fig:architecture}b, our mixed-scale tokenization operates in two stages, applied in both teacher and student branches.
\paragraph{Stage 1: Coarse encoding and region selection.}
The input is first partitioned into $N_c$ non-overlapping coarse tokens of size $32\times32$ pixels to which learnable coarse positional encodings are added. The resulting token sequence $X_c^0$ is processed by a ViT backbone $\mathcal{E}$ with $M$ transformer layers, producing the coarse representation $X_c^M$.
Following DINO~\cite{caron2021emerging}, we use the \texttt{[CLS]}-to-patch attention as a label-free estimate of regional importance. Let $s_\ell\in\mathbb{R}^{N_c}$ denote the attention scores at layer $\ell$. Following CF-ViT~\cite{chen2023cf}, attention maps from intermediate-to-deep layers are aggregated using an exponential moving average,
\begin{equation}
\bar{s}_\ell = \beta \bar{s}_{\ell-1} + (1-\beta)s_\ell,
\qquad \ell=4,\ldots,M,
\label{eq:attention_ema}
\end{equation}
where $\beta$ is the momentum coefficient. The top $\alpha N_c$ tokens according to $\bar{s}_M$ are selected for refinement, where $\alpha$ controls the accuracy--compute trade-off.
\paragraph{Stage 2: Refinement and Multi-scale Fusion.}
Selected regions are refined at higher resolution; the rest keep their coarse tokens.
We partition the coarse set as $S_C = S_{CF} \cup S_{CC}$ into a selected part $S_{CF}$
(refined) and a retained part $S_{CC}$, and let $S_F$ be the fine tokens obtained by
subdividing each selected $32\times32$ token into four $16\times16$ tokens.
Re-initializing fine tokens would discard the context learned during coarse encoding;
instead, we project each parent feature through a lightweight projector $g$ and inject
it into its children, so refinement builds upon the coarse representation~\cite{chen2023cf}:
\begin{equation}
\tilde{X}_r^0 = [\text{CLS}]_r \cup S_{CC} \cup \big(S_F \oplus g(S_{CF})\big).
\end{equation}

Here $[\text{CLS}]_r$ is the refined-stage class token and $\oplus$ adds each parent
coarse feature to its four children after broadcasting to the fine grid. The mixed-scale
sequence has  $N_r = \lfloor(1-\alpha)N_c\rfloor + 4\lceil\alpha N_c\rceil$, spatial tokens and is
encoded by the same ViT backbone, $\tilde{X}_r^M=\mathcal{E}(\tilde{X}_r^0)$, giving the
refined \texttt{[CLS]} output $z_r=\tilde{X}_r^M[0]$.

\subsubsection{Loss Formulation.}

Let $z_c^s=X_c^M[0]$ and $z_r^s=\tilde{X}_r^M[0]$ denote the coarse and refined student representations, with $z_c^t$ and $z_r^t$ the corresponding teacher representations. To encourage representation learning and refinement toward diagnostically relevant regions, we apply standard DINO objective to the refined-stage outputs of teacher and student (Fig.~\ref{fig:architecture}a). Let $P_c^s(x)=h'(z_c^s(x))$, $P_r^s(x)=h'(z_r^s(x))$ denote the coarse and refined student distributions, and $P_c^t(x)$, $P_r^t(x)$ the corresponding teacher distributions. The standard DINO objective is

\begin{equation}
L_{\text{DINO}}
=
\sum_{x_1,x_2}
\mathrm{CE}\!\left(
P_r^t(x_1),
P_r^s(x_2)
\right),
\end{equation}

where $h'(\cdot)=\sigma(h(\cdot))$ denotes the softmax-normalized DINO projection head, and $x_1,x_2$ are two augmented views of the same image. Since the refined representation already contains both coarse and refined tokens, an additional coarse-stage supervision is unnecessary.

Although the DINO objective supervises the final mixed-scale representation, it does not explicitly regulate the relationship between the coarse and refined representations. A natural solution is to align them through a cross-scale KL-divergence objective, as used in CF-ViT~\cite{chen2023cf}. Let $P_c^s(x_i)$ and $P_r^s(x_i)$ denote the coarse and refined student predictions for view $x_i$. The asymmetric objective is

\begin{equation}
\Lklasym=
\sum_{i\in\{1,2\}}
\mathrm{KL}
\!\left(
\mathrm{sg}(P_r^s(x_i))
\,\|\,
P_c^s(x_i)
\right),
\label{eq:kl_asym}
\end{equation}

where $\mathrm{sg}(\cdot)$ denotes the stop-gradient operator. This objective encourages the coarse representation to align with the refined prediction, but places no constraint on what the refined representation should contribute beyond coarse context. Since coarse representations already encode substantial diagnostic information, the refined branch can converge to a redundant copy, limiting the benefit of selective refinement.
To encourage complementary specialization, we instead introduce a \emph{symmetric cross-scale KL} objective on the student outputs,

\begin{equation}
\Lklsym=
\sum_{i\in\{1,2\}}
\mathrm{KL}
\!\left(
\mathrm{sg}(P_r^s(x_i))
\,\|\,
P_c^s(x_i)
\right)
-
\lambda
\,
\mathrm{KL}
\!\left(
\mathrm{sg}(P_c^s(x_i))
\,\|\,
P_r^s(x_i)
\right),
\label{eq:kl_sym}
\end{equation}

where $\lambda=0.5$. The first term preserves semantic alignment between coarse and refined predictions, while the second discourages the refined representation from collapsing to the coarse representation, encouraging it to capture complementary information. The weighting parameter $\lambda$ balances these competing objectives: small values lead to redundant representations, whereas excessively large values weaken the DINO alignment signal.

We formulate cross-scale regularization in the prediction space rather than the embedding space. Since the DINO projection head already outputs probability distributions over prototypes, KL divergence naturally compares coarse and refined predictions without introducing additional contrastive objectives, negative samples, or pairwise similarity computations (e.g., InfoNCE~\cite{InfoNCE}/CLIP~\cite{clip}). This keeps the auxiliary objective consistent with DINO while allowing us to explicitly control the relationship between the two representations through the proposed push--pull formulation.
The symmetric formulation is particularly well-suited to our setting, where the coarse and refined branches are intended to be complementary rather than identical. Together, the DINO objective and symmetric cross-scale regularization encourage refined tokens to preserve global tissue context while contributing additional local morphological information. The overall training objective is, thus defined as, $\mathcal{L}= L_{\text{DINO}} + \Lklsym.$

\section{Experiments}

\subsubsection{Experimental Setup.}
We evaluate slide-level discrimination, survival risk prediction, computational cost, and representation quality across our experiments.

\textbf{Datasets.}
We use two public WSI cohorts. (1) \textbf{CAMELYON16}~\cite{bejnordi2017camelyon} contains 270 training and 129 test breast-cancer WSIs. (2) \textbf{TCGA-NSCLC}~\cite{cooper2018TCGA} contains 1{,}042 lung WSIs (530 LUAD / 512 LUSC) and supports two evaluations: LUAD-vs.-LUSC subtype classification and LUAD survival prediction. We use the patient-stratified Snuffy split~\cite{jafarinia2024snuffy}; after quality filtering, 638 slides (333 LUAD / 305 LUSC) are used for training and 256 slides (115 LUAD / 141 LUSC) form the test set. All splits are patient-disjoint. For survival prediction, we restrict evaluation to the LUAD subset, following standard single-subtype WSI survival benchmarks.

\textbf{Patch extraction and inference modes.}
After background removal, WSIs are tiled into $224\times224$ patches at 20$\times$ magnification. CRAFT and all same-recipe SSL baselines use ViT-S, matching the method's efficiency goal and prior evidence that ViT-S is a strong DINO backbone for pathology~\cite{lunit}. We evaluate two inference modes: (1) \textbf{Coarse-32} uses a global $32\times32$ token grid, yielding 49 tokens at 1.16 GMACs per tile; (2) \textbf{Mixed-16/32} refines the top 50\% high-attention regions into $16\times16$ tokens, yielding 124 tokens at 4.01 GMACs.

\textbf{Downstream Tasks.}
Patch encoders are frozen and evaluated with four MIL aggregators: ABMIL (AB), DSMIL (DS), TransMIL (Trans), and max-pooling (Max). CAMELYON16 uses the official test set with a fixed 5-seed MIL protocol. TCGA-NSCLC follows the DSMIL protocol~\cite{li2021dsmil}: five-fold model selection, fixed test set. Classification performance is reported as AUC. Patch classification uses linear probing on frozen encoder features~\cite{caron2021emerging}, with labels derived from the official CAMELYON16 tumor annotations~\cite{bejnordi2017camelyon}.

For TCGA-LUAD survival, we report concordance index (C-index), the fraction of comparable patient pairs whose predicted risk ordering matches observed survival. Following~\cite{Yang2025}, we use CLAM-Survival on our encoder, whose first fully connected layer maps each encoder's feature dimension to a shared hidden size, controlling for feature-dimension effects across encoders. Published C-indices from~\cite{Yang2025} are used for the six reported baselines; CRAFT and H0-mini are evaluated in our pipeline with the same setup.

\textbf{Comparison and compute protocol.}
Unless otherwise specified, published encoders are run under our evaluation pipeline. External-pretraining rows are treated as foundation-model references; controlled claims are made against same-size or same-recipe baselines. Compute is reported as GMACs for one $224\times224$ forward pass, measured directly where possible; ResNet-50, ViT-B/16, and Swin-T use standard literature estimates. For HIPT, cost is dominated by the cell-level ViT-S/16 ($\approx$4.6 GMACs), while its ViT$_{4096}$/ViT$_{\text{WSI}}$ stages amortize to $<$0.02 GMACs/tile.

\textbf{Statistical and label-free analyses.}
Significance uses Welch's $t$-test over five seeds (CRAFT mixed-$16/32$ vs.\ same-recipe DINO ViT-S, CAMELYON16 DSMIL). To assess $\mathcal{L}_{\text{KL-sym}}$ beyond downstream labels, we probe the teacher's refined \texttt{[CLS]} feature $z_r$ on CAMELYON16 test patches. We report alignment~\cite{wang2020understanding} and three spectral descriptors defined in App.~A.

\textbf{Implementation Details.}
We implement \model{} using ViT-Small within the DINO framework~\cite{caron2021emerging}, training on RTX 4090 GPUs. Pretraining follows a cosine learning rate schedule with AdamW and base LR of \( 5 \times 10^{-4} \). In \model, we set  \( \beta = 0.99 \) and \( \alpha = 0.5 \), resulting in $N_c=49$ and $N_r=124$. For reference, uniform ViT-S/16 uses $N_f=196$ tokens. On CAMELYON16 and TCGA-LUNG, DINO variants are pretrained for 100 and 50 epochs, respectively. Afterwards, each MIL training is performed for 50 epochs.
\subsubsection{Results.} 
We report WSI classification on CAMELYON16 and
TCGA-NSCLC, followed by TCGA-LUAD survival prediction.

\textbf{CAMELYON16 classification.}
\begin{table}[t]
\centering
\caption{WSI classification AUC (\%) on CAMELYON16 (mean\,$\pm$\,std over
5 runs). Among reproduced rows, best \textbf{bold}, second \underline{underlined};
$\dagger$ external pretraining;
Avg = mean over DS/AB/Trans/Max MILs.}
\label{tab:cam16}
\resizebox{\textwidth}{!}{%
\begin{tabular}{l l l ccc c c c r}
\toprule
Method & Encoder & Pretrain & DS & AB & Trans & Max & Avg & \#P & GMACs \\
\midrule
\multicolumn{10}{l}{\textit{Reproduced under our 5-seed MIL protocol}}\\
UNI$^{\dagger}$\cite{chen2023UNI}& ViT-L/16 & DINOv2 & 96.06{\scriptsize$\pm$1.85} & 90.04{\scriptsize$\pm$2.62} & \underline{97.26}{\scriptsize$\pm$0.74} & {97.10}{\scriptsize$\pm$1.19} & 95.12 & 307M & 59.70 \\
Virchow$^{\dagger}$\cite{vorontsov2024virchow}& ViT-H/14 & DINOv2 & 95.00{\scriptsize$\pm$1.74} & 93.95{\scriptsize$\pm$2.60} & 94.90{\scriptsize$\pm$1.68} & 93.59{\scriptsize$\pm$0.63} & 94.36 & 632M & 161.99 \\
CONCH$^{\dagger}$\cite{conch} & ViT-B/16 & iBOT+CoCa & 92.47{\scriptsize$\pm$0.98} & 92.34{\scriptsize$\pm$0.89} & 95.77{\scriptsize$\pm$1.16} & 90.49{\scriptsize$\pm$0.33} & 92.77 & 86M & 	16.87\\
H0-mini$^{\dagger}$\cite{h0mini}& ViT-B/14 & DINOv2 & \underline{98.23}{\scriptsize$\pm$0.81} & \textbf{97.58}{\scriptsize$\pm$0.34} & \textbf{97.87}{\scriptsize$\pm$0.37} & \textbf{99.22}{\scriptsize$\pm$0.00} & \textbf{98.23} & 86M & 22.31 \\
CTransPath$^{\dagger}$\cite{WANG2022102559}& Swin-T & SRCL & 85.62{\scriptsize$\pm$2.34} & 88.08{\scriptsize$\pm$1.95} & 96.31{\scriptsize$\pm$0.56} & 90.45{\scriptsize$\pm$2.22} & 90.12 & 28M & 	4.51\\
Lunit$^{\dagger}$\cite{lunit}& ViT-S/16 & DINO   & 86.44{\scriptsize$\pm$18.5} & 92.96{\scriptsize$\pm$3.17} & 95.98{\scriptsize$\pm$1.51} & 87.82{\scriptsize$\pm$17.9} & 90.80 & 22M & 4.6 \\
DINOv2$^{}$\cite{dinov2}& ViT-S/16 & DINOv2 & 95.75{\scriptsize$\pm$4.83} & 94.78{\scriptsize$\pm$1.33} & 96.27{\scriptsize$\pm$0.94} & 89.52{\scriptsize$\pm$17.6} & 94.08 & 22M & 4.6 \\
DINO$^{}$\cite{caron2021emerging}& ViT-S/16 & DINO   & {96.21}{\scriptsize$\pm$0.30} & {96.26}{\scriptsize$\pm$0.80} & 95.69{\scriptsize$\pm$1.47} & 95.95{\scriptsize$\pm$0.98} & {96.03} & 22M & 4.6 \\
\rowcolor{gray!12}
CRAFT coarse-32$^{}$ & ViT-S/16 & DINO   & 94.89{\scriptsize$\pm$0.62} & 95.36{\scriptsize$\pm$0.90} & 93.19{\scriptsize$\pm$0.59} & 95.21{\scriptsize$\pm$0.71} & 94.66 & 22M & \textbf{1.16} \\
\rowcolor{gray!12}
CRAFT mixed-16/32$^{}$ & ViT-S/16 & DINO   & \textbf{98.43}{\scriptsize$\pm$0.93} & \underline{96.29}{\scriptsize$\pm$1.36} & {96.71}{\scriptsize$\pm$0.88} & \underline{97.18}{\scriptsize$\pm$0.74} & \underline{97.15} & \textbf{22M} & \underline{4.01} \\
\bottomrule
\end{tabular}}
\label{main_comparisons}
\end{table}
Table~\ref{tab:cam16} evaluates frozen patch encoders on the official split using four MIL aggregators. This setting tests whether the encoder improvement is stable across slide-level heads.

The primary control is the same-recipe DINO ViT-S baseline, which matches the
backbone, pretraining data, and MIL protocol, thus isolating the effect of pretraining objective. CRAFT mixed-16/32 improves DSMIL
by +2.22 AUC points ($p=0.004$, Welch's $t$-test) and maintains AUC above
96\% for all four aggregators. Since the MIL heads are trained on frozen patch
features, this cross-aggregator consistency indicates an encoder-level gain
rather than an interaction with a particular slide classifier.

This improvement is obtained under a lower inference budget than uniform
ViT-S/16. Mixed-16/32 evaluates 124 tokens per tile instead of 196, reducing
compute from 4.6 to 4.01 GMACs while improving the matched DINO baseline.
Coarse-32 provides a more aggressive operating point, reducing compute to
1.16 GMACs and still reaching 94.66 mean AUC. Owing to cross-scale alignment during training, coarse tokens remain discriminative, enabling competitive performance with significantly reduced computation. Thus, CRAFT improves the
controlled high-accuracy setting and, with the same trained encoder, provides a
substantially cheaper inference mode when throughput is prioritized.

The foundation-model rows contextualize this trade-off. Although CRAFT uses no
external pretraining and retains a 22M ViT-S backbone, mixed-16/32 achieves the
second-best average AUC in the table and the best DSMIL AUC. Relative to
H0-mini, the only method with higher average AUC, CRAFT is +0.20 AUC higher on
DSMIL and within 1.1 AUC points on average while requiring 5.6$\times$ fewer
GMACs per tile. It also uses 14.9$\times$ fewer GMACs than UNI and
40.4$\times$ fewer than Virchow. These results place CRAFT on a favorable
accuracy--compute frontier: near-foundation-model CAMELYON16 performance with a
compact encoder and single-digit tile-level GMACs.

\textbf{TCGA-NSCLC subtyping.}
\begin{table}[t]
\centering
\small
\setlength{\tabcolsep}{4.5pt}
\caption{WSI subtype classification on TCGA-Lung (LUAD vs. LUSC), AUC (\%) on the fixed 256-slide test set, averaged over four MIL aggregators. Each aggregator value is the mean over five model-selection folds. \emph{Top:} public encoders trained on massive data; \emph{bottom:} low-data 22M encoders. Best \textbf{bold}, second \underline{underlined} among 22M encoders. $\P$\,pretraining overlaps our TCGA test split, so AUC may be inflated.}
\label{tab:tcga_lung}
\resizebox{0.8\textwidth}{!}{%
\begin{tabular}{@{\hspace{1.2em}}c l l l c r@{}}
\toprule
& Method & Encoder & Pretraining & Mean AUC & \#P \\
\midrule
\multicolumn{5}{@{}l}{\textit{Frozen public Encoder trained on Massive Data}}\\
\multirow{6}{*}{\rotatebox[origin=c]{90}{\textit{Massive Data}}}
& UNI~\cite{chen2023UNI}         & ViT-L/16 & DINOv2, Mass-100K      & 96.81 & 307M \\
& Virchow~\cite{vorontsov2024virchow} & ViT-H/14 & DINOv2, MSK-1.5M   & 96.67 & 632M \\
& H0-mini$^{\P}$~\cite{h0mini}   & ViT-B/14 & DINOv2 distill, TCGA & 97.79 & 86M \\
& CTransPath$^{\P}$~\cite{WANG2022102559} & Swin-T & SRCL, TCGA+PAIP & 92.81 & 28M \\
& RetCCL$^{\P}$     & ResNet-50 & TCGA+PAIP (CCL)  & $85.61$ & 23.5M \\
& Lunit$^{\P}$~\cite{lunit}      & ViT-S/16 & DINO, TCGA           & 94.13 & 22M \\
\cmidrule(l){1-6}
\multirow{7}{*}{\rotatebox[origin=c]{90}{\textit{Similar-scale Data}}}
& Barlow Twins \cite{mammadov2025} \cite{barlow} & ViT-S/16 & Barlow, TCGA-LUNG & 91.7 & 22M \\
& MoCo v3   \cite{mammadov2025}\cite{moco3}   & ViT-S/16 & MoCo v3, TCGA-LUNG & 95.0 & 22M \\
& DINO      \cite{mammadov2025} \cite{caron2021emerging}   & ViT-S/16 & DINO, TCGA-LUNG   & 95.0 & 22M \\
& DINO ViT-S            & ViT-S/16 & DINO, CAM16          & 85.32 & 22M \\
& IN ViT-S~\cite{dosovitskiy2021imageworth16x16words,steiner2022trainvitdata} & ViT-S/16 & sup., IN-21k$\to$1k & 84.08 & 22M \\
& \cellcolor{gray!15}Ours coarse-32$^{}$     & \cellcolor{gray!15}ViT-S/16 & \cellcolor{gray!15}DINO, TCGA-LUNG & \cellcolor{gray!15}\underline{95.89} & \cellcolor{gray!15}22M \\
& \cellcolor{gray!15}Ours mixed-16/32$^{}$ & \cellcolor{gray!15}ViT-S/16 & \cellcolor{gray!15}DINO, TCGA-LUNG & \cellcolor{gray!15} \textbf{96.24} & \cellcolor{gray!15}\textbf{22M} \\
\bottomrule
\end{tabular}}
\end{table}
Table~\ref{tab:tcga_lung} reports WSI subtyping on TCGA-NSCLC (LUAD vs.\ LUSC), averaged over four MIL aggregators; per-aggregator results with std are in App.~C. This tests whether the mixed-scale encoder generalizes beyond CAMELYON16 to a different organ and task. Among comparable 22M encoders, CRAFT mixed-16/32 obtains
the highest mean AUC, exceeding MoCov3/DINO~\cite{mammadov2025} by +1.24 points and CAMELYON16-pretrained DINO transfer baseline by
+10.92 points. The two CRAFT inference modes then separate the role of adaptive
refinement: coarse-32 already reaches 95.89 AUC at 1.16 GMACs per tile, while
mixed-16/32 raises performance to 96.24 AUC at 4.01 GMACs. The consistent
per-aggregator gains in App.~C indicate that refined tokens add
subtype-discriminative information without degrading coarse representation.

The foundation-model comparison shows how far this compact setting can be pushed
against models trained at much larger scale. CRAFT is within 0.6 AUC of UNI and
Virchow while using a 22M backbone and requiring 14.9$\times$ and 40.4$\times$
fewer GMACs per tile, respectively. It also exceeds Lunit by +2.11 points and
CTransPath by +3.43 points; these rows are marked with $\P$ because their
pretraining includes TCGA slides and may overlap the fixed test distribution.
H0-mini remains the strongest row, but combines a larger ViT-B encoder,
distillation from a billion-scale teacher, and the same TCGA-overlap caveat.
Taken together, the matched 22M rows establish the controlled gain, while the
foundation-model rows show that CRAFT approaches large-scale pretrained
encoders on TCGA-NSCLC at substantially lower tile-level compute.
\begin{figure}[t]
    \centering
    \vspace{0pt}
    \includegraphics[width=0.6\linewidth]{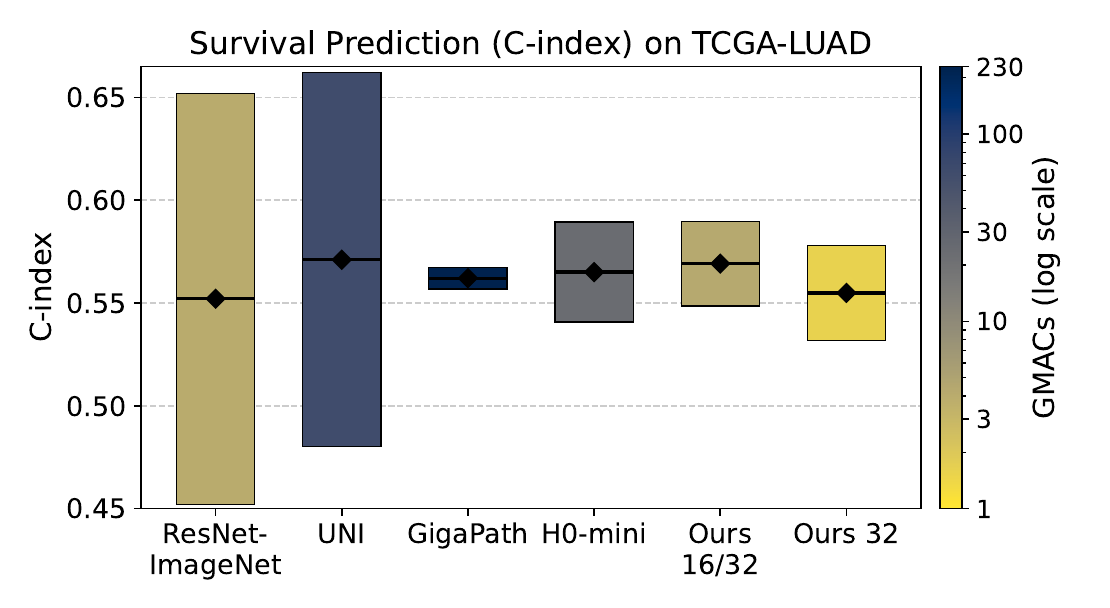}
    \caption{Survival prediction on TCGA-LUAD (selected encoders). Bars show mean\,$\pm$\,std across 5 folds; color encodes compute (GMACs).}
    \label{fig:survival_luad}
\end{figure}
\textbf{TCGA-LUAD survival.}
Fig.~\ref{fig:survival_luad} compares TCGA-LUAD C-index; full values are in
App.~D. Survival prediction evaluates whether the frozen
encoder transfers from diagnostic discrimination to patient-level risk ordering
under a fixed survival head. CRAFT mixed-16/32 attains 56.9$\pm$2.06 C-index, exceeding H0-mini by +0.4 points, GigaPath by
+0.7 points, and trailing only UNI by 0.2 points. It also has lower cross-fold
variance than the other top-scoring methods except GigaPath. This performance
requires only 4.01 GMACs per tile: about 5.9$\times$ less than H0-mini, over
15$\times$ less than UNI, and over 50$\times$ less than GigaPath. Coarse-32
reaches 55.49$\pm$2.30 C-index at 1.16 GMACs, preserving a low-compute survival
setting.

To further assess representation quality, we evaluate driver-gene mutation prediction on TCGA-LUAD (App.~B). Despite using a compact ViT-S backbone, CRAFT remains competitive with pathology foundation models up to $7\times$ larger.

\begin{table*}
\begin{minipage}[t]{0.59\textwidth}
\centering

\captionof{table}{Ablations on CAMELYON16. MIL metrics average four aggregators (DS, AB, Trans, Max). \textbf{Best} and \underline{second-best} marked. TS: Token Sizes; MF: Multi-scale Fusion.}
\label{tab:results_ablation}

\resizebox{\linewidth}{!}{
\begin{tabular}{c l c c c c c c c c c c c}
    \toprule
    & & & & & & \multicolumn{3}{c}{\textbf{Linear Probing}} &
    \multicolumn{3}{c}{\textbf{MIL}} \\
    \cmidrule(lr){7-9}
    \cmidrule(lr){10-11}
    ID & Model & TS & MF & $\Lklasym$ & $\Lklsym$
    & ACC & Prec & Recall & Avg AUC & Avg Recall  & GMACs \\
    \midrule

    B1 & Baseline & 16 & -- & -- & -- & 97.35 & 90.8 & 69.0 & 96.03 & 95.61 & 4.6 \\
    B2 & Baseline & 32 & -- & -- & -- & 98.05 & 89.2 & \textbf{81.4} & 94.59 & 93.68 & 1.16 \\

    \midrule
    \multicolumn{12}{l}{\textit{CRAFT operating points}} \\

    O1 & \model & 16 & \checkmark & \checkmark & $\times$
    & \underline{98.16} & 92.3 & \underline{80.7}
    & \underline{96.74} & \underline{96.74}
    & 5.76 \\

    O2 & \model & 16/32 & \checkmark & \checkmark & $\times$
    & 97.80 & \textbf{93.8} & 73.2
    & 95.65 & 94.19
    & 4.01 \\

    O3 & \model & 16/32 & \checkmark & $\times$ & \checkmark
    & \textbf{98.62} & \underline{93.2} & 77.8
    & \textbf{97.15} & \textbf{97.25}
    & 4.01 \\

    \midrule
    \multicolumn{12}{l}{\textit{Coarse-branch operating points}} \\
    
    E1 & \model & 32 & \checkmark & \checkmark & $\times$
    & 97.85 & 93.1 & 74.7
    & 94.83 & 94.47
    & 1.16 \\

    E2 & \model & 32 & \checkmark & $\times$ & \checkmark
    & 97.98 & 93.4 & 76.5
    & 94.66 & 92.45
    & 1.16 \\

    \bottomrule
\end{tabular}
}
\end{minipage}
\hfill
\begin{minipage}[t]{0.38\textwidth}
\centering

\captionof{table}{Label-free representation analysis on CAMELYON16 test
($\Lklasym$ vs.\ $\Lklsym$). Mean$\pm$std over 5 runs; best \textbf{bold}.}
\label{tab:repr-effect}

\setlength{\tabcolsep}{2.pt}

\resizebox{\linewidth}{!}{
\begin{tabular}{lcc}
\toprule
Metric &
$\mathcal{L}_{\text{KL-asym}}$ &
$\mathcal{L}_{\text{KL-sym}}$ \\
\midrule


Alignment $\downarrow$
& $0.234{\pm}0.003$
& $\mathbf{0.225{\pm}0.003}$ \\

RankMe $\uparrow$
& $263.0{\pm}0.8$
& $\mathbf{264.4{\pm}0.8}$ \\

Effective rank $\uparrow$
& $275.4{\pm}0.7$
& $\mathbf{279.7{\pm}0.7}$ \\

$\alpha$-ReQ ($\to 1$)
& $1.61{\pm}0.00$
& $\mathbf{1.48{\pm}0.00}$ \\

\midrule

Max-MIL AUC $\uparrow$
& $95.7$
& $\mathbf{97.2}$ \\

\bottomrule
\end{tabular}
}

\end{minipage}

\end{table*}

\subsubsection{Ablation Studies.} Table~\ref{tab:results_ablation} evaluates the contributions of coarse-to-fine tokenization, multi-scale fusion, and cross-scale regularization. Linear probing measures tile-level representation quality, while MIL evaluates the transfer of frozen features to slide-level prediction. The ablations support three main conclusions.

First, fine-scale representations benefit from incorporating coarse contextual
information. The coarse-only baseline (B2) outperforms the fine-only baseline (B1)
on tile-level recall (81.4\% vs.\ 69.0\%) yet underperforms at slide level,
indicating it lacks the fine detail needed for robust WSI classification. The
uniformly refined CRAFT variant (O1) instead improves on \emph{both}: MIL Avg AUC by
$0.71$\,pp over B1 ($96.03\to96.74$) and tile-level recall by $0.11$\,pp over B2,
despite the same $16\times16$ token resolution. Combining coarse context with fine
detail thus yields a small but cross-task-robust gain.

Second, selectively allocating fine resolution is more effective than uniformly refining every region, but only when coupled with proposed $\Lklsym$.

Compared with uniform refinement (O1), refining only the most salient half of the image (O2) reduces computation from 5.76 to 4.01 GMACs, but also degrades downstream performance when trained with the asymmetric cross-scale objective. Replacing the asymmetric objective with the proposed symmetric formulation (O2$\rightarrow$O3) recovers and further improves performance, increasing MIL Avg AUC from 95.65 to 97.15 at the same computational cost. Notably, O3 outperforms the more expensive uniformly refined model (O1), showing that selectively allocating spatial resolution is more effective than uniformly increasing it. 

Finally, $\Lklasym$ encourages redundancy: coarse-only
features (E1) outperform mixed-scale (O2) in linear probing ($97.85$ vs.\ $97.80$).
In contrast, $\Lklsym$ restores complementarity: mixed-scale features (O3) surpass coarse-only (E2) in both LP ($98.62$ vs.\ $97.98$) and MIL
($97.15$ vs.\ $94.66$). Interestingly, $\Lklsym$ also improves
coarse-only features ($97.98$ vs.\ $97.85$), indicating that enforcing complementary
specialization yields better-structured representations at both scales, not only
when combined. 
This supports our design: the symmetric term specializes the two
scales rather than collapsing them together. The same trend holds across $\lambda$
(App.~E). A component analysis can be found at App.~F.

\subsubsection{Representation-Level Effect of $\Lklsym$.} As shown in Table~\ref{tab:repr-effect}, the symmetric loss yields more
view-invariant features: alignment, which is the distance between embeddings of two
augmented views, drops ($0.225$ vs.\ $0.234$, a ${\sim}3\sigma$ gap). The three
spectrum-entropy measures (RankMe~\cite{garrido2023rankme}, effective
rank~\cite{effRank}, and the eigenspectrum-decay exponent
$\alpha$~\cite{agrawal2022alpha}, all quantifying how well the embedding space is utilized) favour $\Lklsym$, indicating capacity is used broadly
with no sign of collapse. These entropy- and invariance-based metrics are validated
by downstream predictors (RankMe by design; alignment intrinsic to the SSL objective)
and track the Max-MIL AUC gain, so the symmetric loss redistributes capacity towards
a flatter spectrum and stronger cross-view consistency.

\begin{figure*}[t]
    \centering
    \vspace{-2mm} 
    \begin{subfigure}{0.4\textwidth}
        \centering
        \includegraphics[width=\textwidth]{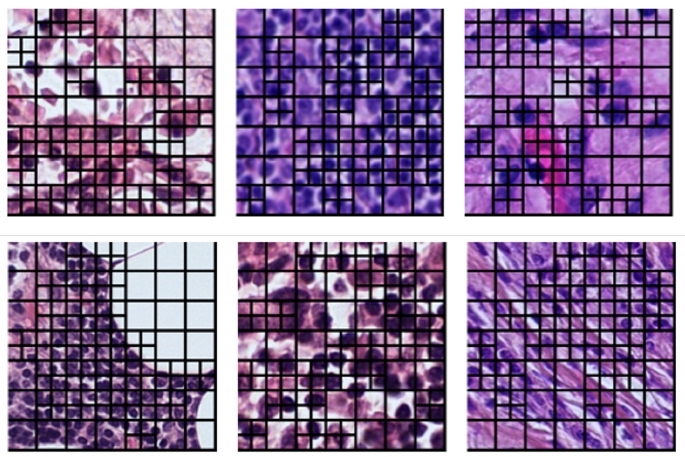}
        \vspace{-4mm} 
        \caption{}\label{fig:qual_a}
    \end{subfigure}
    \hspace{0.03\textwidth}
    \begin{subfigure}{0.45\textwidth}
        \centering
        \vspace{-2mm} 
        \includegraphics[width=\textwidth]{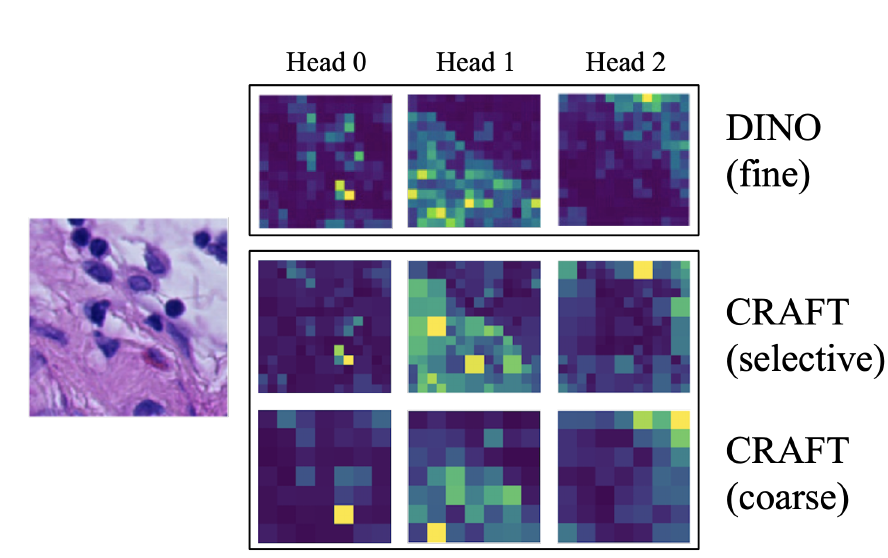}
        \vspace{-4mm} 
        \caption{}\label{fig:qual_b}
    \end{subfigure}
    \vspace{-3mm}
    \caption{
    \textbf{(a)} Mixed-scale tokenization on CAMELYON16 patches: key regions are
    refined to fine tokens while background and stroma stay coarse. \textbf{(b)}
    Self-attention maps for selected heads, comparing DINO (fine) with \model{} under
    selective and coarse inference; input shown left.}
    \label{fig:qualitative}
\end{figure*}

\subsubsection{Visual Analysis.}
Fig.~\ref{fig:qualitative} (a) illustrates the adaptive mixed-scale tokenization: high-attention regions with dense cellular structures are refined into fine-resolution tokens, while background and stromal regions remain coarse. Fig.~\ref{fig:qualitative} (b) shows self-attention maps from selected heads. \model{} (mixed-scale) closely matches DINO (fine), confirming that mixed-scale tokenization preserves diagnostic features. Even under coarse-only inference, similar attention patterns persist, demonstrating effective multi-scale distillation.

\section{Conclusion}
We introduced CRAFT, a self-supervised framework that learns \emph{where} to allocate spatial resolution during DINO pretraining, selectively refining salient regions while preserving coarse context. A symmetric cross-scale KL objective encourages complementary coarse and fine representations. Across multiple pathology tasks, CRAFT surpasses comparable-scale SSL baselines and remains competitive with substantially larger pathology foundation models at much lower inference cost, despite using a compact 22M-parameter backbone trained on modest data. These results suggest that selective spatial resolution allocation is an efficient alternative to uniform tokenization. Future work includes replacing the fixed refinement ratio $\alpha$ with an image-adaptive budget and extending the symmetric objective to patch-level distillation (e.g., iBOT), where masked patch targets remain misaligned across scales.

\section*{Acknowledgements}
The work described in this paper was conducted in the framework of the Graduate School 2543/2 “Intraoperative Multi-Sensory Tissue Differentiation in Oncology" (project ID 40947457) funded by the German Research Foundation (DFG - Deutsche Forschungsgemeinschaft). This work has been supported by the Deutsche Forschungsgemeinschaft (DFG) – EXC number 2064/1 – Project number 390727645 and SFB 1233, TP 1, Project number 276693517. The authors thank the International Max Planck Research School for Intelligent Systems (IMPRS-IS) for supporting A. Stammer, V. Bundele and M. Hosseinzadeh.

%
%

%
%
%
%
\bibliographystyle{splncs04}
\bibliography{source/egbib}

\input{083-supp/083-supp}

\end{document}

%% file: 083-supp/083-supp.tex
\renewcommand{\thesubsection}{\Alph{subsection}}
\onecolumn 
\section*{Appendix}
\label{sec:appendix}


\subsection{Label-free metrics}
\label{app:metrics}
All metrics use the frozen refined-stage \texttt{[CLS]} features. Arrows in the table of the main paper mark the
preferred direction. \emph{Alignment}~\cite{wang2020understanding}: mean squared
distance between $\ell_2$-normalised embeddings of two views (lower $=$ more
view-invariant). \emph{RankMe}~\cite{garrido2023rankme}: entropy-based
smooth effective rank of the feature singular values. \emph{Effective
rank}~\cite{effRank}: exponential Shannon entropy of the normalised singular values
(higher $=$ flatter spectrum). \emph{$\alpha$-ReQ}~\cite{agrawal2022alpha}: power-law
exponent of eigenspectrum decay ($\alpha\!\approx\!1$ well-conditioned).
\emph{Participation ratio}~\cite{partRatio}:
$(\sum_i\lambda_i)^2/\sum_i\lambda_i^2$, effective dimensionality dominated by the
leading eigenvalues.

\subsection{Driver-Gene Mutation Prediction on TCGA-LUAD}
\label{app:mutation}
\paragraph{Task.} As a further probe of representation quality, we evaluate
weakly-supervised prediction of somatic driver-gene mutation status from H\&E
WSIs of lung adenocarcinoma (LUAD)~\cite{coudray2018}. We treat four of the
driver genes highlighted by Coudray et al.~\cite{coudray2018} for LUAD
histology---\emph{KRAS}, \emph{EGFR}, \emph{TP53}, \emph{STK11}---as independent
binary slide-level tasks (mutated vs.\ wild-type).

\paragraph{Dataset and labels.} We use the diagnostic slides of the TCGA-LUAD
cohort. Mutation calls are taken from the cBioPortal~\cite{cerami2012,gao2013}
PanCancer Atlas file; following standard practice we retain only non-silent
mutations and reduce them to a patient-level binary label per gene.

\paragraph{Protocol.} As for other tasks in the main paper, encoders are frozen and only the DSMIL~\cite{li2021dsmil} aggregator is trained, so the encoder is the only variable. Encoders include Lunit~\cite{lunit}, UNI~\cite{chen2023UNI}, Virchow~\cite{vorontsov2024virchow}, H0-mini~\cite{h0mini}, and CRAFT. For each gene, we train a single-logit MIL classifier for 50 epochs under patient-level 5-fold stratified CV (seed 42; no patient spans train/test). We report mean\,$\pm$\,std over the 5 folds, following~\cite{wong2023milfeatsel}.

\begin{table}[!]
\centering
\caption{Driver-gene mutation prediction on TCGA-LUAD. Weakly-supervised ROC-AUC (\%), DSMIL~\cite{li2021dsmil}
  aggregation; encoder is the only variable. Mean\,$\pm$\,std over 5 patient-level,
  label-stratified folds; Macro = mean over genes. \textbf{Bold} and \underline{Underlined} show the first and second best methods per column.}
\label{tab:mutation}
\begin{tabular}{llccccc}
\toprule
Encoder & Arch.\ (dim) & KRAS & EGFR & TP53 & STK11 & Macro \\
\midrule
Virchow  & ViT-H/14 (2560) & \ms{54.3}{5.9} & \ms{54.0}{4.3} & \ms{60.6}{5.8} & \ms{49.7}{11.0} & 54.65 \\
H0-mini  & ViT-B/14 (1536) & \ms{\underline{56.4}}{5.1} & \ms{54.6}{7.6} & \ms{\textbf{69.4}}{6.2} & \ms{\textbf{64.1}}{6.7} & \textbf{61.13} \\
UNI      & ViT-L/16 (1024) & \ms{56.3}{4.7} & \ms{\underline{55.6}}{3.7} & \ms{63.7}{5.1} & \ms{55.2}{5.4} & 57.71 \\
\midrule
CRAFT-mixed  & ViT-S/16 (384)  & \ms{\textbf{57.8}}{4.5} & \ms{\textbf{56.8}}{6.0} & \ms{\underline{65.5}}{3.4} & \ms{\underline{61.3}}{8.9} & \underline{60.35} \\
\bottomrule
\end{tabular}
\end{table}

Driver-mutation status is partly reflected in H\&E morphology and can be inferred under weak supervision, though signal strength varies by gene and drops out-of-domain~\cite{coudray2018}. We therefore read this as a representation probe rather than a clinical result. Despite its ViT-S backbone, CRAFT is competitive with foundation models up to $7\times$ larger, leading on KRAS and EGFR. TCGA slides originate from many submitting sites with distinct staining and scanning signatures; models can exploit these rather than biology, inflating absolute AUCs across all encoders~\cite{howard2021}. H0-mini's edge on TP53/STK11 likely reflects distillation from a much larger teacher rather than same-size training; moreover, if this distillation used TCGA, H0-mini would also carry data leakage.

\subsection{TCGA-NSCLC: Per-Aggregator Results}
\label{app:lung}
Table~\ref{tab:app_tcga_lung} expands the main-text mean AUC
(Table~2) into per-aggregator scores. Model selection uses
5-fold CV on the 638 training slides, with evaluation applied on the fixed 256-slide test
set. Public encoders are frozen and evaluated on our split under our MIL
protocol. Barlow Twins, MoCo v3, and DINO results are quoted from Mammadov et
al.~\cite{mammadov2025}. 
Encoders marked $\P$ are pretrained on TCGA, overlapping our test split, so their
AUC is potentially inflated (H0-mini, Lunit, CTransPath, RetCCL); UNI and Virchow training distribution doesn't contain TCGA.

Among the 22M encoders, CRAFT mixed-16/32 is the only method above 94 AUC for all
four aggregators (94.4--97.7). Same-size baselines are far less stable under the
weakest aggregator: Max-pooling drops Lunit to 89.80$\pm$5.69 and the
CAMELYON16-pretrained DINO baseline to 68.58$\pm$6.50, whereas CRAFT retains
95.34$\pm$0.41. Since all rows share the frozen-encoder, fixed-aggregator
protocol, this consistency reflects encoder quality rather than aggregator tuning.

\begin{table}[!]
\centering
\small
\setlength{\tabcolsep}{4.5pt}
\caption{Per-aggregator WSI subtype classification on TCGA-NSCLC (LUAD vs.\ LUSC),
AUC (\%). DS/AB/Trans/Max = DSMIL, ABMIL, TransMIL, Max-pooling; Avg = mean of the
four. Among 22M encoders (\emph{Low Data}), best \textbf{bold}, second
\underline{underlined}. $\P$\,TCGA overlap may inflate AUC.}
\label{tab:app_tcga_lung}
\resizebox{\textwidth}{!}{%
\begin{tabular}{@{\hspace{1.2em}}c l l l ccccc r}
\toprule
& Method & Encoder & Pretraining & DS & AB & Trans & Max & Avg & \#P \\
\midrule
\multicolumn{10}{@{}l}{\textit{Frozen public Encoder trained on Massive Data}}\\
\multirow{6}{*}{\rotatebox[origin=c]{90}{\textit{Massive Data}}}
& UNI~\cite{chen2023UNI}         & ViT-L/16 & DINOv2, Mass-100K      & \ms{97.82}{0.49} & \ms{95.47}{1.87} & \ms{98.10}{1.15} & \ms{95.83}{0.34} & 96.81 & 307M \\
& Virchow~\cite{vorontsov2024virchow} & ViT-H/14 & DINOv2, MSK-1.5M   & \ms{97.53}{0.21} & \ms{{95.37}}{1.12} & \ms{97.80}{1.37} & \ms{95.96}{0.28} & 96.67 & 632M \\
& CTransPath$^{\P}$~\cite{WANG2022102559} & Swin-T & SRCL, TCGA+PAIP & \ms{93.80}{0.28} & \ms{91.21}{2.20} & $96.88_{\pm0.99}$ & \ms{89.34}{0.46} & 92.81 & 28M \\
& RetCCL$^{\P}$     & ResNet-50 & TCGA+PAIP (CCL)  & $87.80_{\pm1.26}$ & $83.57_{\pm1.38}$ & $95.14_{\pm1.97}$ & $75.92_{\pm2.11}$ & $85.61$ & 23.5M \\
& H0-mini$^{\P}$~\cite{h0mini}   & ViT-B/14 & DINOv2 distill, TCGA & \ms{98.35}{0.19} & \ms{96.90}{1.10} & \ms{98.18}{0.90} & \ms{97.71}{0.24} & 97.79 & 86M \\
& Lunit$^{\P}$~\cite{lunit}      & ViT-S/16 & DINO, TCGA           & \ms{95.13}{0.32} & \ms{93.45}{1.58} & \ms{98.12}{0.76} & \ms{89.80}{5.69} & 94.13 & 22M \\
\cmidrule(l){1-10}
\multirow{7}{*}{\rotatebox[origin=c]{90}{\textit{Low Data}}}
& Barlow Twins \cite{mammadov2025} \cite{barlow} & ViT-S/16 & Barlow, TCGA-LUNG & 86.0 & 91.7 & 94.4 & 94.6 & 91.7 & 22M \\
& MoCo v3   \cite{mammadov2025}\cite{moco3}   & ViT-S/16 & MoCo v3, TCGA-LUNG & 92.3 & \textbf{95.2} & 95.7 & 96.9 & 95.0 & 22M \\
& DINO      \cite{mammadov2025} \cite{caron2021emerging}   & ViT-S/16 & DINO, TCGA-LUNG   & 93.0 & 94.1 & 96.1 & 96.8 & 95.0 & 22M \\
& DINO ViT-S            & ViT-S/16 & DINO, CAM16          & \ms{91.40}{1.43} & \ms{84.87}{1.71} & \ms{96.41}{1.02} & \ms{68.58}{6.50} & 85.32 & 22M \\
& IN ViT-S~\cite{dosovitskiy2021imageworth16x16words,steiner2022trainvitdata} & ViT-S/16 & sup., IN-21k$\to$1k & \ms{85.89}{1.21} & \ms{84.86}{7.01} & \ms{94.96}{1.38} & \ms{70.62}{5.50} & 84.08 & 22M \\
& \cellcolor{gray!15}Ours coarse-32$^{}$      & \cellcolor{gray!15}ViT-S/16 & \cellcolor{gray!15}DINO, TCGA-LUNG & \cellcolor{gray!15}\ms{\underline{97.36}}{0.42} & \cellcolor{gray!15}\ms{{94.06}}{2.15} & \cellcolor{gray!15}\ms{\underline{97.49}}{0.98} & \cellcolor{gray!15}\ms{\underline{94.65}}{0.16} & \cellcolor{gray!15}\underline{95.89} & \cellcolor{gray!15}22M \\
& \cellcolor{gray!15}Ours mixed-16/32$^{}$   & \cellcolor{gray!15}ViT-S/16 & \cellcolor{gray!15}DINO, TCGA-LUNG & \cellcolor{gray!15}\ms{\textbf{97.65}}{0.45} & \cellcolor{gray!15}\ms{\underline{94.44}}{1.97} & \cellcolor{gray!15}\ms{\textbf{97.54}}{0.93} & \cellcolor{gray!15}\ms{\textbf{95.34}}{0.41} & \cellcolor{gray!15}\textbf{96.24} & \cellcolor{gray!15}22M \\
\bottomrule
\end{tabular}
}
\end{table}

\subsection{Survival Prediction}
\label{app:survival}
Table~\ref{tab:survival_luad} reports C-index on TCGA-LUAD using the fixed CLAMSurvival aggregator~\cite{Yang2025} with 5-fold cross-validation. Results for all competing encoders are taken directly from~\cite{Yang2025}, while CRAFT is evaluated under the identical protocol. Absolute C-indices lie between 0.51 and 0.57, reflecting the inherent difficulty of TCGA-LUAD survival prediction due to its limited cohort size, heavy censoring, and weak slide-level prognostic signal rather than the evaluation setup. Similar performance ranges have been reported in prior work under different survival pipelines (e.g., DS-MIL achieves 0.537~\cite{li2021dsmil}).

Using a common survival head isolates differences in the learned representations. Notably, encoders with nearly identical computational cost can exhibit substantially different survival performance (e.g., HIPT: 53.8 vs.\ CTransPath: 51.2 at $\sim$4.5 GMACs), indicating that representation quality, rather than computational budget, is the dominant factor. While a survival head individually tuned to each encoder could improve absolute performance, such tuning would confound representation quality with downstream optimization. We therefore adopt a fixed aggregator to enable a controlled comparison across encoders.

\begin{table}[!]
    \centering
    \caption{Survival prediction (C-index) on TCGA-LUAD. $\dagger$External pretraining. All methods use CLAMSurvival~\cite{Yang2025}; C-indices except Ours are quoted from~\cite{Yang2025}. GMACs: single $224\times224$ forward pass (ResNet-50/ViT-B/Swin-T from literature; HIPT dominated by its cell-level ViT-S/16, $\approx$4.6).}
    \begin{tabular}{lllc}
    \toprule
    Model & Encoder & C-index & GMACs \\
    \midrule
    DINO-HistoPretrain$^\dagger$~\cite{Yang2025} & ViT-B/16 & 56.9 $\pm 5.5$ & ~17.6 \\
    GigaPath$^\dagger$~\cite{Xu2024} & ViT-G/14 & $56.2 \pm 0.52$ & 228.1 \\
    UNI$^\dagger$~\cite{chen2023UNI} & ViT-L/16 & {$\textbf{57.1} \pm 9.1$} & 61.55 \\
    H0-mini$^\dagger$~\cite{h0mini} & ViT-B/14 & $56.5 \pm 2.43$ & ~23.5 \\    
    HIPT$^\dagger$~\cite{chen2022HIPT} & Hier.\ ViT & $53.8 \pm 4.0$ & ~4.6\\
    CTransPath$^\dagger$~\cite{WANG2022102559} & Swin-T & $51.2 \pm 5.0$ & ~4.5 \\
    ResNet-ImageNet$^\dagger$~\cite{Yang2025} & ResNet-50 & $55.2 \pm 10.0$ & ~4.1 \\
    
    \midrule
    Ours coarse-32$^{}$  & ViT-S/16 & $55.49 \pm  2.30$ & \textbf{1.16} \\ 
   Ours mixed-16/32$^{}$  & ViT-S/16 & $\underline{56.9} \pm 2.06$ & \underline{4.01} \\  
    \bottomrule
    \end{tabular}%
    \label{tab:survival_luad}
\end{table}

\subsection{$\lambda$-Sensitivity of the Symmetric-KL Loss on NCT-CRC-HE-100K}
\label{app:lambda}

To isolate the effect of the symmetric-KL weight $\lambda$ (Eq.5), we sweep it on NCT-CRC-HE-100K\cite{Kather2019}, a nine-class colorectal tile-classification benchmark of 100k $224\times224$ H\&E tiles ($20\times$). We use two disjoint evaluation sets. In-domain, we split the NCT tiles into an 80k training partition and a 20k held-out validation partition. Externally, we use the patient-disjoint CRC-VAL-HE-7K test set ($7{,}180$ tiles, 50 patients), which is touched only for final evaluation.

We pretrain CRAFT (ViT-S/16, $\alpha=0.5$) for 100 epochs on the 80k training tiles (self-supervised; labels unused), varying only $\lambda$ and holding the rest of the recipe fixed. Following the linear-probing protocol of the main paper, we then freeze the teacher and fit a logistic-regression head on the mixed-scale (refined) \texttt{[CLS]} features of the 80k labeled training tiles, selecting the probe's regularization strength on the 20k validation partition (reported as ``Val.\ Bal.\ Acc.''). We report the resulting balanced accuracy on the external CRC-VAL-HE-7K set. As label-free measures of representational capacity, we also report the effective rank~\cite{effRank} and participation ratio~\cite{partRatio} of the \texttt{[CLS]} features — soft counts of how many dimensions they span (higher = more spread, less collapsed).

We fix $\lambda=0.5$, the default in our main experiments; among the reported settings it also attains the best accuracy on the held-out NCT validation split. At this weight, CRAFT reaches $90.97$ balanced accuracy on the external test set using $124$ tokens and $4.0$ GMACs. Its \texttt{[CLS]} features span a high-dimensional subspace, with effective rank $13.3$ and participation ratio $6.7$.

To probe the effect of the weight, we report larger values. Increasing $\lambda$
drives the coarse- and fine-stage head distributions apart—the coarse--fine KL grows
from $2\times10^{-4}$ at $\lambda=0.5$ to $5\times10^{-2}$ at $\lambda=5$—but this
divergence is detrimental rather than beneficial: balanced accuracy declines mildly
at $\lambda=2$ (89.64) and collapses at $\lambda=5$ (77.87), with the effective rank
and participation ratio falling in step (to 6.1 and 3.6). An excessively large
$\lambda$ over-regularizes the two scales and degrades the representation; the small
default $\lambda=0.5$ stays well clear of this regime.

\begin{table}[!]
\centering
\caption{$\lambda$-sensitivity on NCT-CRC-HE-100K (mixed-scale \texttt{[CLS]}; single run per row).
Row shading marks the regime: \colorbox{green!12}{safe} $\to$ \colorbox{orange!16}{caution} $\to$
\colorbox{red!18}{over-regularized collapse}. cls\_sim/KL are model-level (final epoch).
$^{*}$Operating point selected on validation Bal.\ Acc.}
\label{tab:lambda}
\resizebox{\textwidth}{!}{
\begin{tabular}{l cc c cc cc}
\toprule
 & \multicolumn{2}{c}{\textbf{Coarse--fine}} & \textbf{Val.}
 & \multicolumn{2}{c}{\textbf{Test (mixed-scale \texttt{[CLS]})}}
 & \multicolumn{2}{c}{\textbf{Label-free}} \\
 
\cmidrule(lr){2-3}\cmidrule(lr){4-4}\cmidrule(lr){5-6}\cmidrule(lr){7-8}
\textbf{$\lambda$ (KL$_{\text{sym}}$)} & cls\_sim & KL ($\times10^{-3}$) & Bal.\ Acc.
& Bal.\ Acc. & Macro-F1 & Eff.\ rank $\uparrow$ & Part.\ ratio $\uparrow$ \\
\midrule

\rowcolor{green!12}  0.5 (default)   & 0.993 & 0.16   & \textbf{98.81} & \textbf{90.97} & \textbf{90.46} & \textbf{13.3} & \textbf{6.7} \\

\rowcolor{orange!16} 2.0             & 0.975 & 0.35   & 98.09 & 89.64 & 89.55 & 10.4 & 5.5 \\
\rowcolor{red!18}    5.0             & 0.824 & 54.3 & 89.47 & 77.87 & 77.32 & 6.1  & 3.6  \\
\bottomrule
\end{tabular}
}
\end{table}


\begin{figure}[!t]
\centering
\begin{subfigure}[t]{0.49\textwidth}
  \centering
  \includegraphics[width=\textwidth]{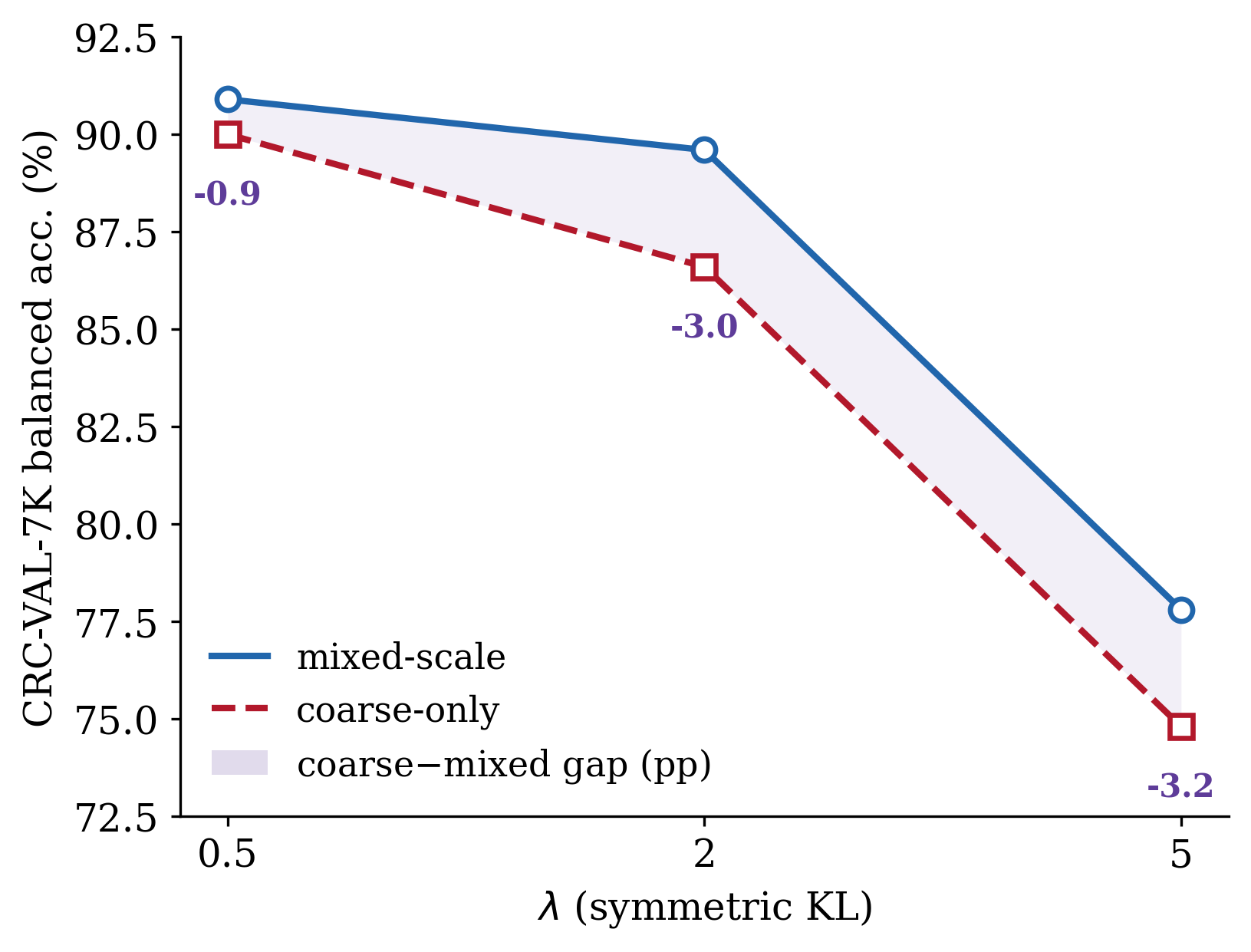}
  \caption{Coarse-only vs.\ mixed-scale probe accuracy; the coarse$-$mixed gap
  widens monotonically with $\lambda$ ($-0.9\to-3.2$\,pp).}
  \label{fig:coarse_gap}
\end{subfigure}
\hfill
\begin{subfigure}[t]{0.49\textwidth}
  \centering
  \includegraphics[width=\textwidth]{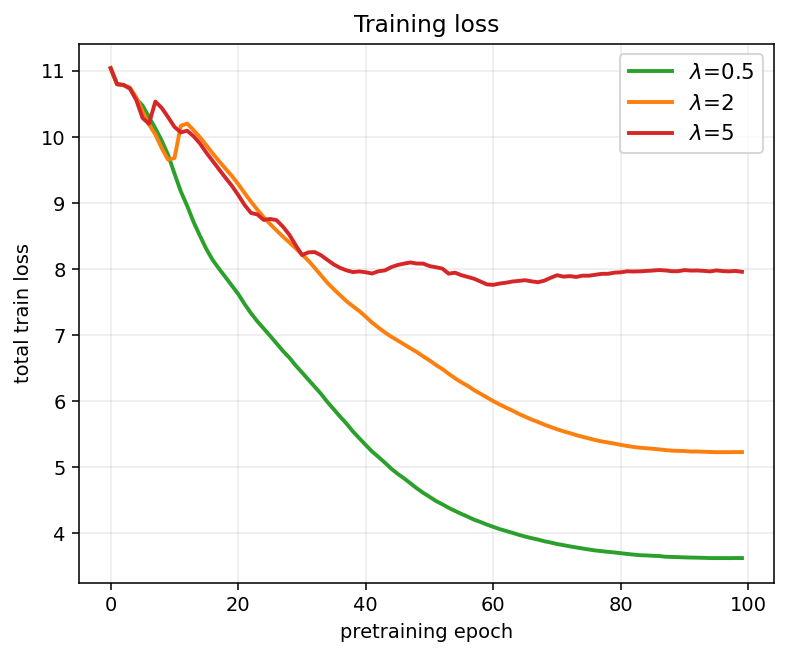}
  \caption{Training dynamics: alignment holds for $\lambda\le1$, destabilizes at
  $\lambda=5$; loss converges higher and stalls (${\approx}8$).}
  \label{fig:lambda_dynamics}
\end{subfigure}
\caption{$\lambda$-sensitivity of CRAFT (ViT-S/16; single run per $\lambda$) on
NCT-CRC-HE-100K / CRC-VAL-HE-7K.}
\label{fig:lambda_all}
\end{figure}

\paragraph{The coarse stage pinpoints where $\lambda$ acts first.}
Mixed-scale accuracy and the coarse--fine KL hold up well through $\lambda{=}2$ and
collapse only at $\lambda{=}5$, whereas the coarse-only representation degrades
earlier and monotonically (Fig.~\ref{fig:coarse_gap}). Coarse and mixed inference
nearly coincide at $\lambda{=}0.5$; as $\lambda$ grows, the coarse$-$mixed gap widens
monotonically ($-0.9\to-3.0\to-3.2$\,pp). Over-regularization thus hits the coarse
stream first: coarse tokens rely on cross-scale alignment to stay discriminative and
lose it before the mixed representation is affected.

\paragraph{Training dynamics.}
The training-loss trajectories show how optimization fails as $\lambda$ grows (Fig.~\ref{fig:lambda_dynamics}). At $\lambda{=}0.5$ the loss converges smoothly (${\approx}3.6$); at $\lambda{=}2$ optimization slows and plateaus higher (${\approx}5.2$); at $\lambda{=}5$ training destabilizes early (${\sim}$epoch~30) and stalls near $8.0$.


These trajectories mirror the alignment in
Table~\ref{tab:lambda}: the symmetric term opposes coarse--fine \texttt{[CLS]}
alignment more strongly as $\lambda$ grows, so
stronger repulsion yields a higher, unconverged loss. This tug-of-war between
cross-scale alignment and symmetric repulsion explains the collapse at large
$\lambda$; the default $\lambda{=}0.5$ leaves optimization essentially undisturbed.

\subsection{Component Ablation}
\label{app:ablation}

Table~\ref{tab:ablation} isolates each component of \model{}during the adaptive selection, on CAMELYON16, on the frozen-feature
tile-classification (Linear Probing) and slide-level (MIL) protocols. \textbf{TS}
denotes the token scale(s) used at inference ($16$, $32$, or mixed $16/32$),
\textbf{MF} multi-scale fusion, and $\Lklasym$/$\Lklsym$ the asymmetric and symmetric
cross-scale alignment losses. Best and second-best per column are \textbf{bold} and
\underline{underlined}; MIL Max is reported over five seeds ($\pm$\,std).

\begin{table}[!]
\centering
\caption{Component ablation of \model. Each block removes one component from a
\model{} operating point (full $\to$ ablated); $\Delta$ is the change on the
key metric of that block. LP Recall is tumor recall.}
\label{tab:ablation}
\resizebox{\linewidth}{!}{
\begin{tabular}{c l c c c c c c c c c c c}
    \toprule
    & & & & & & \multicolumn{3}{c}{\textbf{Linear Probing}} &
    \multicolumn{3}{c}{\textbf{MIL}} \\
    \cmidrule(lr){7-9}\cmidrule(lr){10-12}
    ID & Model & TS & MF & $\Lklasym$ & $\Lklsym$
    & ACC & Prec & Recall & Avg AUC & Avg Recall & Max & GMACs \\
    \midrule
    O1 & \model & 16 & \checkmark & \checkmark & $\times$
    & 98.16 & 92.3 & 80.7 & 96.74 & 96.74 & 97.67{\scriptsize$\pm$0.48} & 5.76 \\
    C1 & \model & 16 & $\times$ & \checkmark & $\times$
    & 97.97 & 94.2 & 75.6 & 95.59 & 94.80 & 95.66{\scriptsize$\pm$1.18} & 5.76 \\
    \midrule
    \multicolumn{13}{l}{\textit{MF effect (mixed, $16/32$)}} \\
    O2 & \model & 16/32 & \checkmark & \checkmark & $\times$
    & 97.80 & 93.8 & 73.2 & 95.65 & 94.19 & 95.66{\scriptsize$\pm$2.76} & 4.01 \\
    C2 & \model & 16/32 & $\times$ & \checkmark & $\times$
    & 97.73 & 92.9 & 73.0 & 92.76 & 91.02 & 94.37{\scriptsize$\pm$0.63} & 4.01 \\
    \midrule
    \multicolumn{13}{l}{\textit{$\Lklasym$ effect (mixed, $16/32$)}} \\
    O2 & \model & 16/32 & \checkmark & \checkmark & $\times$
    & 97.80 & 93.8 & 73.2 & 95.65 & 94.19 & 95.66{\scriptsize$\pm$2.76} & 4.01 \\
    C3 & \model & 16/32 & \checkmark & $\times$ & $\times$
    & 97.60 & 92.9 & 71.0 & 93.77 & 89.79 & 94.23{\scriptsize$\pm$0.01} & 4.01 \\
    \bottomrule
\end{tabular}
}
\end{table}

We isolate multi-scale fusion (MF) and cross-scale alignment ($\Lklasym$). For fine
tokens ($16$), removing MF (O1$\to$C1) cuts tumor recall by $5.1$\,pp
($80.7\to75.6$), confirming the value of reusing coarse context. For mixed tokens
($16/32$), the effect shifts to slide level: removing MF (O2$\to$C2) drops MIL Avg
AUC by $2.9$\,pp ($95.65\to92.76$), indicating coarse context matters most when fine
tokens are only partially available. Finally, removing $\Lklasym$ (O2$\to$C3) causes
the largest MIL Avg-Recall drop ($4.4$\,pp, $94.19\to89.79$), showing cross-scale
alignment is necessary for effective slide-level classification.

\subsection{Attention Head Outputs}
We further examine the self-attention maps of the six heads in the final teacher layer, complementing the main-text analysis.

\begin{figure}[!]
    \centering
    \includegraphics[width=\textwidth]{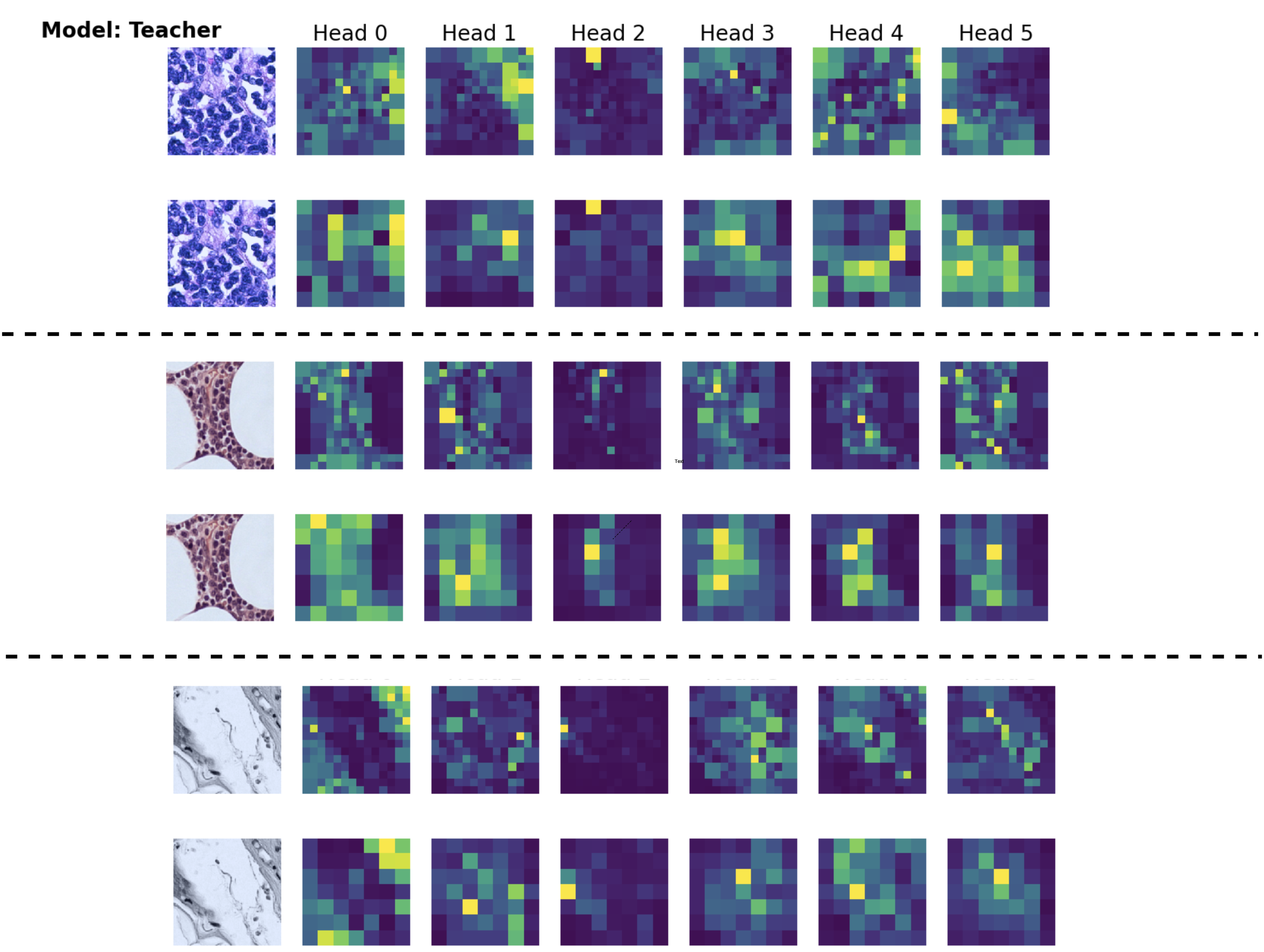}
    \caption{Self-attention maps from the six final-layer teacher heads for three tissue regions (input on the left). Top row of each pair: selective grid; bottom row: coarse grid. Brighter is higher attention.}
    \label{fig:head_out}
\end{figure}

Figure~\ref{fig:head_out} shows three tissue regions, each under the selective (top) and coarse (bottom) grid. The heads divide labor: different heads light up on different parts of the patch rather than all tracking the same region. More importantly, the coarse grid (bottom rows) reproduces the attention structure of the selective grid (top rows) closely—the same regions remain salient at lower resolution. This is the visual counterpart of our quantitative finding that coarse-only inference stays discriminative.

\subsection{Mixed-Scale Grids}
\begin{figure}[!]  
    \centering 
    \includegraphics[width=0.6\textwidth]{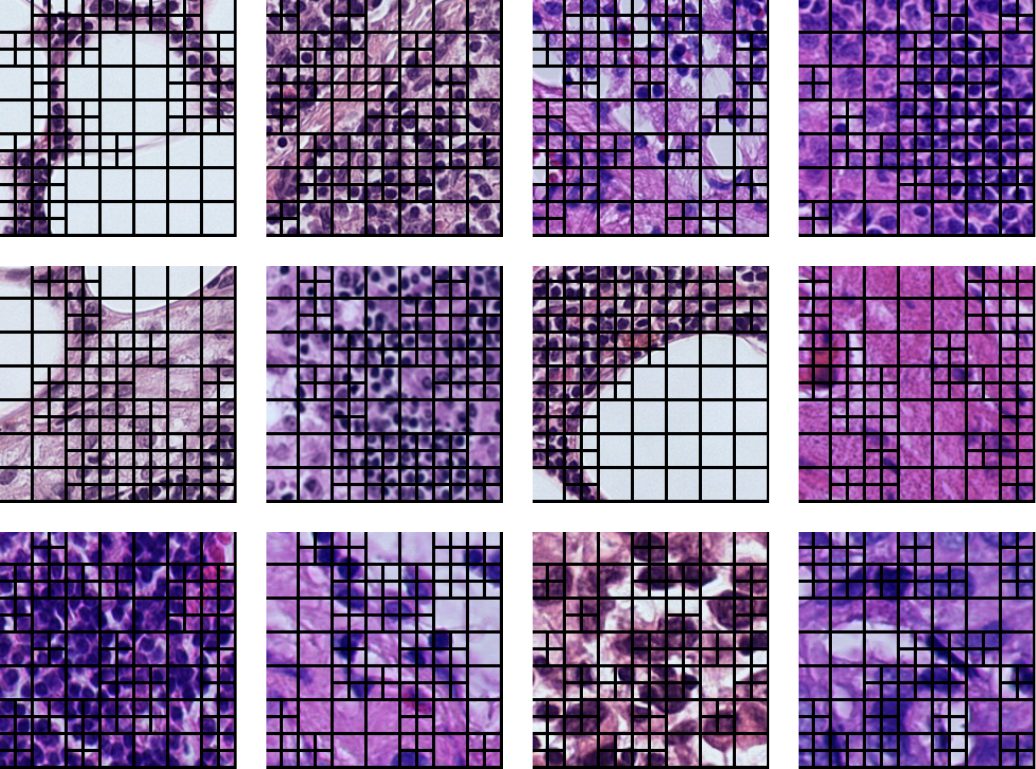}  
    \caption{Visualization of adaptive mixed-scale tokenization applied to histopathology patches. High-attention regions, typically rich in cellular structures, are refined into fine-grained tokens, while background and less informative stromal areas remain at a coarse resolution. This hierarchical tokenization strategy preserves diagnostically relevant morphological features while optimizing computational efficiency.} 
    \label{fig:mixed_scale_grids} 
\end{figure}

Figure~\ref{fig:mixed_scale_grids} shows additional mixed-scale tokenization examples. Refinement consistently concentrates on cellular regions, while stromal and background tissue stay coarse, showing that learned selection is spatially selective rather than uniform.

\subsection{Student--Teacher Comparison}
\label{sec:student_teacher}
\begin{figure}[!t]
    \centering
    \includegraphics[width=0.6\textwidth]{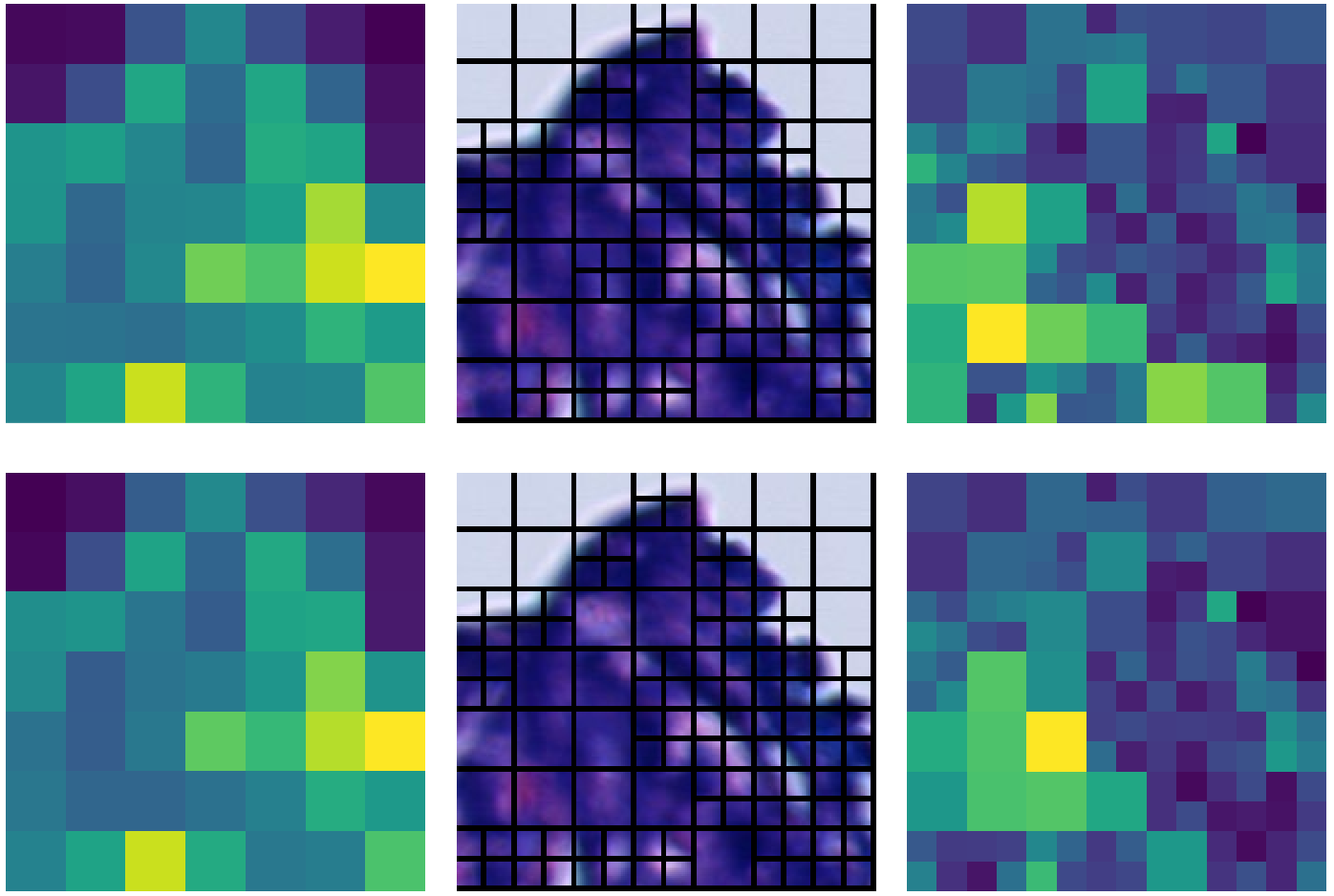}
    \caption{\textbf{Teacher--student selection agreement.}
    Top: teacher branch; bottom: student branch. \emph{Left:} global
    CLS attention. \emph{Center:} selected tokens overlaid on the input.
    \emph{Right:} CLS attention. Both branches attend to and select the
    same regions, showing the two-branch design preserves a consistent
    token selection.}
    \label{fig:branch}
\end{figure}

Figure~\ref{fig:branch} compares the \texttt{[CLS]} attention and token
selection of the teacher and student branches on the same input. The two
branches converge to a near-identical selection: the global \texttt{[CLS]}
attention (left), the selected-token grid overlaid on the input (center), and
the per-layer \texttt{[CLS]} attention (right) all coincide across the two rows,
concentrating on the same salient regions.

This agreement indicates that the two-branch design does not push the branches
toward divergent selections, and that the added adaptivity does not destabilize
optimization (final training loss 1.93 vs.\
1.26 for the baseline). Unlike SimPrune~\cite{simprune}, where single-branch selection prunes the two
branches inconsistently, our branches converge to the same selection without an
explicit cross-branch mechanism.

\subsection{Limitations}
\begin{figure}[!]  
    \centering 
    \includegraphics[width=0.6\textwidth]{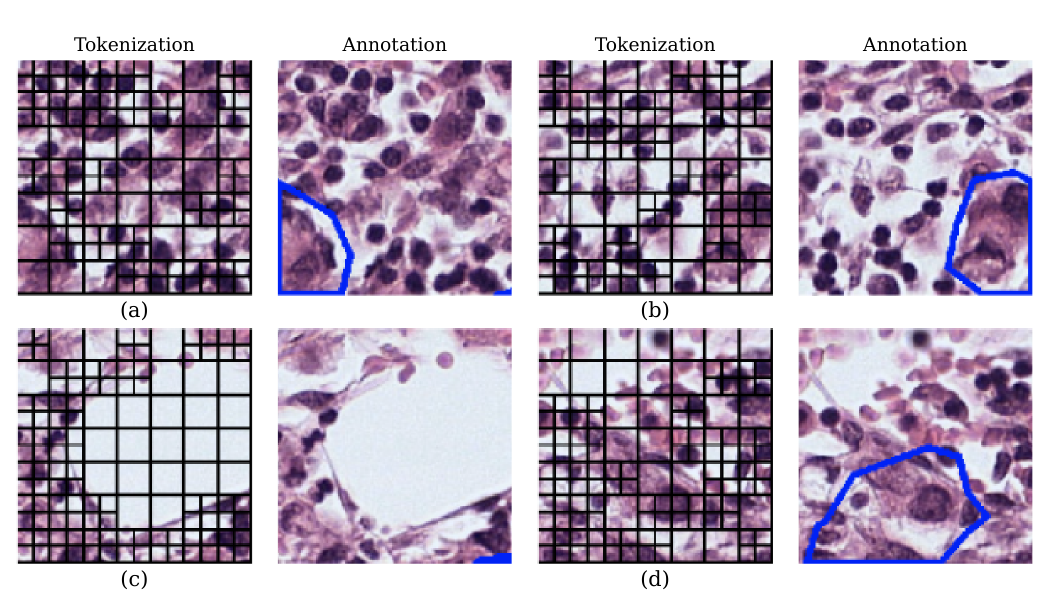}  
    \caption{Mixed-scale tokenization vs.\ tumor annotations on CAMELYON16.
Each pair (a--d): CRAFT tokenization grid (left) and pathologist annotation of
tumor outlined in blue (right). Finer tokens follow dense/high-contrast morphology, not the annotated tumor extent.}
    \label{fig:limit} 
\end{figure}

CRAFT has some limitations. First, although the learned refinement policy consistently improves downstream performance, the selected regions are not explicitly optimized for diagnostic interpretability and often follow general morphological structure rather than clinically salient tissue (Fig.~\ref{fig:limit}). Second, CRAFT uses a fixed refinement ratio ($\alpha=0.5$), whereas adapting the refinement budget to image complexity could further improve the accuracy--efficiency trade-off. Finally, while we evaluate CRAFT across multiple pathology tasks, broader validation on additional cohorts and imaging modalities is needed to establish its generality.

